\documentclass[journal]{IEEEtran}
\usepackage
{
    graphicx,
    amssymb,
    amsmath,
    amsthm,
    xcolor,
    dsfont,
    algpseudocode,
    authblk,
}

\usepackage[ngerman, english]{babel}
\usepackage{bm}
\usepackage{enumitem}
\usepackage{algorithm}
\usepackage{tabularx}
\usepackage[colorlinks,
            pdffitwindow=false,
            plainpages=false,
            pdfpagelabels=true,
            pdfpagemode=UseOutlines,
            pdfpagelayout=SinglePage,
            bookmarks=false,
            colorlinks=true,
            hyperfootnotes=false,
            linkcolor=blue,
            citecolor=blue]
{hyperref}

\usepackage{todonotes}

\newcommand{\R}{\mathbb{R}}

\newcommand{\Acal}{\mathcal{A}}
\newcommand{\Dcal}{\mathcal{D}}
\newcommand{\Scal}{\mathcal{S}}
\newcommand{\Tcal}{\mathcal{T}}
\newcommand{\Xcal}{\mathcal{X}}
\newcommand{\Zcal}{\mathcal{Z}}

\newcommand{\E}{\mathbb{E}}

\newcommand{\Pset}{\mathcal{P}}
\newcommand{\Pfront}{\mathcal{P}_{\mathcal{F}}}

\newcommand{\norm}[1]{\left\|#1\right\|}

\newcommand{\condSet}[2]{\left\{ #1 ~ \middle\vert~ #2 \right\}}

\newtheorem{remark}{Remark}

\begin{document}
%
\title{Multi-objective Deep Learning: Taxonomy and Survey of the State of the Art}
%
%
\author{Sebastian Peitz and Sedjro Salomon Hotegni
\thanks{Both authors are with the Department of Computer Science, TU Dortmund,
Dortmund, Germany, and with the Lamarr Institute for Machine Learning and Artificial Intelligence, e-mail: \{sebastian.peitz, salomon.hotegni\}@tu-dortmund.de.}
}

%

\maketitle

\begin{abstract}
Simultaneously considering multiple objectives in machine learning has been a popular approach for several decades, with various benefits for multi-task learning, the consideration of secondary goals such as sparsity, or multicriteria hyperparameter tuning. However---as multi-objective optimization is significantly more costly than single-objective optimization---the recent focus on deep learning architectures poses considerable additional challenges due to the very large number of parameters, strong nonlinearities and stochasticity. This survey covers recent advancements in the area of multi-objective deep learning. We introduce a taxonomy of existing methods---based on the type of training algorithm as well as the decision maker's needs---before listing recent advancements, and also successful applications. All three main learning paradigms supervised learning, unsupervised learning and reinforcement learning are covered, and we also address the recently very popular area of generative modeling.
\end{abstract}


\section{Introduction}
Conflicting design or decision criteria are everywhere, and the challenge of identifying suitable decisions has been around for a very long time. To address this problem, we are looking for the set of optimal compromises---also called the \emph{Pareto set} after the Italian economist Vilfredo Pareto. In simple terms, a decision is Pareto optimal if we are unable to further improve all objectives at the same time, but instead have to accept a trade-off if we want to further improve one of the criteria. In mathematical terms, this can be cast as a \emph{multi-objective optimization problem}, see \cite{Miettinen1998,Ehr05} for detailed introductions. Problems of this type have been studied in a large number of fields, examples being political decision-making \cite{Swamy2023}, medical therapy planning \cite{Kuefer2003}, the design of numerical algorithms \cite{Breugel2020}, or the control of complex dynamical systems \cite{PD18a}.

Very naturally, multiple criteria also arise in the area of machine learning, for instance if we want to solve multiple tasks with a single model, which results in several performance measures. Alternatively, we might be interested in secondary objectives such as sparsity/efficiency, robustness or interpretability. Specifically in the area of reinforcement learning, researchers have made strong arguments to intensify multi-objective research \cite{roijers2015multi,Vamplew2022}. However, since both multi-objective optimization and deep learning can be computationally expensive, their combination gives rise to particular challenges in terms of algorithmic efficiency and decision-making.

The goal of this survey paper is to provide a detailed overview of the state of the art in multi-objective deep learning. A large number of articles has appeared in recent years, and progress has been rapid. To keep an overview and ease the entry for researchers interested in---but until now less familiar with---multi-objective learning, we provide a taxonomy of the various approaches (Section \ref{sec:Taxonomy}), before surveying the current state of the art in Section \ref{sec:Survey}. 
We consider the three main learning paradigms supervised learning (\ref{subsec:MODL}), unsupervised learning (\ref{subsec:DUL}) and reinforcement learning (\ref{sec:MORL}---since this is a special case, we will discuss it separately), and we also provide an overview of generative modeling (\ref{subsec:GenAI}), neural architecture search (\ref{subsec:NAS}) and successful applications of multi-objective deep learning (\ref{subsec:Applications}). Finally, we briefly comment on the vice-versa combination in Section \ref{sec:DLforMO}, i.e., the usage of deep learning to enhance the solution of multi-objective optimization problems.

\section{Preliminaries}\label{sec:Preliminaries}
Before introducing our taxonomy and survey to multi-objective deep learning we here give brief introductions to deep learning (\ref{subsec:DL}) and multi-objective optimization (\ref{subsec:MO}), respectively and highlight the specifics of multi-objective machine learning in Section \ref{subsec:MOML}.

\subsection{Deep learning}\label{subsec:DL}
This section's main purpose is to introduce notation and to highlight the similarity between various areas of machine learning when it comes to training and optimization. We briefly cover supervised (\ref{subsubsec:SupervisedDL}) and unsupervised learning (\ref{subsubsec:UnsupervisedLearning}) as well as generative modeling (\ref{subsubsec:Generative}). For much more detailed introductions, the reader is referred to one of the excellent text books \cite{Bishop2006,Hastie2009,GBC16,Bishop2024}.

\subsubsection{Supervised learning}\label{subsubsec:SupervisedDL}
The overarching theme in supervised learning is function approximation from examples. In other words, we want to approximate an unknown input-to-output mapping $y=f(x)$, $f:\R^n \rightarrow \R^m$, by a parametrized function $f_\theta$. 
Here, $\theta\in\R^q$ represents the (usually high-dimensional) vector of trainable parameters. There exist many options how to construct $f_\theta$, such as polynomials, Fourier series, or Gaussian processes. The specialty in deep learning \cite{lecun2015deep} is that $f_\theta$ represents a deep neural network with $\ell$ layers, in which affine transformations $(W,b)$ alternate with nonlinear activation functions $\sigma$:
\begin{align*}
    z^{(0)} &= x, \\
    z^{(j)} &= \sigma^{(j)}\left(W^{(j)} z^{(j-1)} + b^{(j)} \right), \quad j = 1, \ldots, \ell, \\
    y & =z^{(\ell)}.
\end{align*}
All trainable weights are collected in $\theta = \left\{\left(W^{(j)},b^{(j)}\right)\right\}_{j=1}^\ell$.

\begin{remark}
    Besides the just-mentioned classical \emph{feed-forward} architecture, there are numerous other deep learning models used for function approximation such as \emph{convolutional networks}, \emph{residual networks}, or \emph{recurrent networks} with feedback loops \cite{GBC16,Bishop2024}. However, the goal of this article is not to cover the details of deep learning models, but to shed light on the training with multiple criteria, which is why we are not going into more detail here.
\end{remark}

To fit $f_\theta$ to $f$, we use a dataset $\Dcal = \{(x_i,y_i)\}_{i=1}^N$ of $N$ samples and minimize the \emph{empirical loss} $L$:
\begin{align}
    \theta^* = \arg\min_{\theta \in \R^q} L(\theta). \label{eq:DLtraining}
\end{align}
A common choice for $L$ is the mean squared error, i.e.,
\[ 
    L(\theta) = \frac{1}{N}\sum_{i=1}^N \norm{y_i - f_\theta(x_i)}_2^2,
\]
but there are numerous alternatives as well as additional terms (e.g., for regularization or sparsity \cite{Lemhadri2021}). Regardless of the specific choice, Problem \eqref{eq:DLtraining} is a high-dimensional, nonlinear optimization problem. The typical approach to find $\theta^*$ (or at least a $\theta$ that yields a satisfactory performance) is gradient-based optimization using backpropagation, often in combination with stochasticity and momentum, as in the popular Adam algorithm \cite{KB14} or adaptations thereof (e.g., \cite{Bungert2022}). Consequently, the gradient $\nabla L(\theta)$ plays a central role in training.

\subsubsection{Unsupervised and self-supervised learning}
\label{subsubsec:UnsupervisedLearning}

Unlike supervised learning, which depends on labeled data to model input-output relationships, unsupervised and self-supervised learning focus on extracting meaningful representations or patterns from unlabeled data \cite{9442775, sindhu2020survey}. While unsupervised learning relies on directly identifying latent structures in the data \cite{chen2022semi, 9952910, wilson2020survey}, self-supervised learning uses surrogate tasks, derived from the data itself, to create labels and train models that can generalize to downstream tasks \cite{gui2023survey, khan2024survey}.

The objectives in unsupervised learning often involve clustering \cite{caron2018deep}, dimensionality reduction \cite{saul2003think}, or density estimation \cite{wang2019nonparametric}, formulated as optimization problems. Clustering focuses on partitioning a dataset $\Dcal = \{(x_i)\}_{i=1}^N$ into $N_C$ clusters. A popular objective is to minimize intra-cluster variance while maximizing inter-cluster separation:
\[
 \min_{C_1, \dots, C_K} \sum_{k=1}^{N_C} \sum_{x_i \in C_k} \|x_i - \mu_k\|^2, \quad \mu_k = \frac{1}{|C_k|} \sum_{x_i \in C_k} x_i,
\]
where $C_k$ is the set of data points assigned to the $k^{\text{th}}$ cluster, and $\mu_k$ is its centroid. Techniques like K-means and Gaussian Mixture Models (GMMs) \cite{reynolds2009gaussian} address this optimization. 
Modern neural clustering approaches embed data into latent spaces, enhancing the flexibility and scalability of clustering methods \cite{yang2019deep, li2022neural}. For dimensionality reduction, the goal is to find a lower-dimensional representation for high-dimensional data, retaining as much information as possible \cite{jia2022feature}. Autoencoders, in particular, leverage neural networks to learn compressed representations by minimizing a reconstruction loss \cite{michelucci2022introduction, berahmand2024autoencoders}, e.g., 
\[
    \min_{\theta,\phi} \sum_{i=1}^N \norm{x_i - g_{\phi}(f_{\theta}(x_i))}_2^2,
\]
where $f_{\theta}(x)=z$ encodes the input into a latent variable $z\in\R^m$, and $g_{\phi}(z)=\tilde{x}$ decodes the latent variable back to an output $\tilde{x}$ that is equal in size to the input. If we choose a small latent space dimension $m$ and successfully optimize for a small loss (i.e., $g_{\phi}(f_{\theta}(x_i))=\tilde{x}_i \approx x_i$), then we have successfully found an intrinsic, low-dimensional structure that encodes the information of the data set.
In density estimation, probabilistic methods like Kernel Density Estimation (KDE) \cite{chen2017tutorial} and Variational Autoencoders (VAEs) \cite{Kingma2013,kingma2019introduction} model the underlying distribution of data. These methods provide insights into data regularities and outliers, making them useful for anomaly detection and data generation \cite{nachman2020anomaly, bustos2023ad} (see also the next Section \ref{subsubsec:Generative} for details).

Self-supervised learning trains models by creating surrogate tasks, transforming unlabeled data into structured problems resembling supervised learning. These tasks encourage the model to learn representations that generalize well to downstream tasks. Common pretext tasks include contrastive learning and generative models \cite{liu2021self, kumar2022contrastive}. Self-supervised learning employs predictive tasks as well. Notable examples include predicting the next frame in a video \cite{jang2024visual}, determining the rotation angle of an image \cite{jing2018self}, and reconstructing masked portions of input data \cite{he2022masked}. Masked Autoencoders (MAEs) \cite{he2022masked}, for example, reconstruct masked inputs $x$ by minimizing:
$${L}_\text{MAE} = \|x_\text{masked} - g_\theta(f_\phi(x_\text{visible}))\|_2^2,$$
where $f_\phi$ is the encoder that processes the visible patches $x_\text{visible}$ to produce latent representations, and $g_\theta$ is the decoder that reconstructs the masked input $x_\text{masked}$ from these representations.
Problems in both unsupervised and self-supervised deep learning are commonly addressed through gradient-based optimization methods, where a loss function quantifying the task objective (e.g., reconstruction error, contrastive loss) is minimized using backpropagation and stochastic gradient descent or its variants.


\subsubsection{Generative modeling}
\label{subsubsec:Generative}
The central goal of generative modeling (cf.\ \cite{Ruthotto2021} for a very clear and concise introduction) is to learn an unknown probability distribution $\Xcal$ using training samples $x \sim \Xcal$, the task is thus closely related to the self-supervised learning framework described in the previous section. To this end, one constructs a generative model $g:\R^p \rightarrow \R^n$ that maps points $z \in \R^p$---drawn from a lower-dimensional, more tractable probability distribution $\Zcal$---to $x$ in such a manner that $g(z) \sim \Xcal$.
As an example, consider a generator $g$ that maps points $z$ drawn from a multivariate Gaussian distribution to images $x$ in such a way that the generated images follow the same probability distribution as the training data. In other words, the set of generated images is statistically indistinguishable from the set of real images, since $p(g(z)) = p(x)$.
Since the introduction of Generative Adversarial Networks (GANs) in 2014 by Goodfellow et al.\ \cite{Goodfellow2014} and of variational autoencoders by Kingma and Welling in 2013 \cite{Kingma2013} (see also the previous section \ref{subsubsec:UnsupervisedLearning}), this area of machine learning has gained massive attention, with a large number of new architectures such as diffusion models \cite{Sohl-Dickstein2015,rombach2022high}, and culminating in the recent wave of transformer-based architectures \cite{Vaswani2017} and large language models (LLMs).

To draw the connection to gradient-based learning, let's briefly consider the GAN architecture, which consists of a generator $g_{\theta}:\R^p\rightarrow\R^n$ and a discriminator $f_\phi:\R^n \rightarrow [0,1]$, both of which can be realized by deep neural networks. The task of the generator is to transform inputs $z\sim\Zcal$ into outputs $x$, while the discriminator's job is to decide whether an input $x$ has been drawn from a real dataset, or was created by the generator. It is thus a standard binary classifier predicting the probability that $x$ is real. As a consequence, the two networks share the same loss function
\[
    \resizebox{\linewidth}{!}{$L_{\mathsf{GAN}}(\theta,\phi) = \E_{x\sim\Xcal}\left[ \log(f_\phi(x)) \right] + \E_{z\sim\Zcal}\left[ \log(1 - f_\phi(g_\theta(z))) \right].$}
\]
The specialty is that the two networks are \emph{adversarials}, meaning that the generator seeks to minimize $L_{\mathsf{GAN}}$ (i.e., to fool the discriminator), while the discriminator wants to maximize $L_{\mathsf{GAN}}$ (that is, to perform correct classifications).
Training thus usually alternates between gradient descent of $L_{\mathsf{GAN}}$ using the gradient with respect to $\theta$, and gradient ascent of $L_{\mathsf{GAN}}$ using the gradient with respect to $\phi$, i.e.,
\begin{align*}
    \theta^{(i+1)} &= \theta^{(i)} - \eta_\theta\left(\theta^{(i)}\right) \nabla_\theta L_{\mathsf{GAN}}\left(\theta^{(i)},\phi^{(i)}\right), \\ 
    \phi^{(i+1)} &= \phi^{(i)} + \eta_\phi\left(\phi^{(i)}\right) \nabla_\phi L_{\mathsf{GAN}}\left(\theta^{(i)},\phi^{(i)}\right),
\end{align*}
where $\eta_\theta$ and $\eta_\phi$ are (potentially variable) learning rates.
Training stops when this two-player game converges to an equilibrium.

\subsection{Multi-objective optimization}\label{subsec:MO}

Again, this section covers only the basic concepts of multi-objective optimization. Much more detailed overviews can be found in \cite{Miettinen1998,Ehr05}.

\subsubsection{Definition and concepts}
Consider the situation where instead of a single loss function, we have a vector with $K$ conflicting ones, i.e., $L(\theta) = [L_1(\theta), \ldots, L_K(\theta)]^\top$. The task thus becomes to minimize all losses at the same time, i.e.,
\begin{equation}\label{eq:MOP}
    \min_{\theta\in\R^q} \begin{pmatrix}L_1(\theta) \\ \vdots \\ L_K(\theta)\end{pmatrix}. \tag{MOP}
\end{equation}
If the objectives are conflicting, then there does not exist a single optimal $\theta^*$ that minimizes all $L_k$. Instead, there exists a \emph{Pareto set} $\Pset$ with optimal trade-offs, i.e., 
\[
\resizebox{\linewidth}{!}{$
\Pset = \condSet{\theta\in\R^q}{\nexists\hat\theta:\begin{array}{ll}
    L_k(\hat\theta) \leq L_k(\theta) & \mbox{for}~k=1,\ldots,K, \\
    L_k(\hat\theta) < L_k(\theta) & \mbox{for at least one }~k
\end{array}}.
$}
\]
In other words, a point $\hat{\theta}$ \emph{dominates} a point $\theta$, if it is at least as good in all $L_k$, while being strictly better with respect to at least one loss. The Pareto set $\Pset$ thus consists of all non-dominated points.
The corresponding set in objective space is called the \emph{Pareto front} $\Pfront=L(\Pset)$. Under smoothness assumptions, both objects have dimension $K-1$ \cite{Hil01}, i.e., $\Pset$ and $\Pfront$ are lines for two objectives, 2D surfaces for three objectives, and so on. Furthermore, they are bounded by Pareto sets and fronts of the next lower number of objectives \cite{GPD19}, meaning that individual minima constrain a two-objective solution, 1D fronts constrain the 2D surface of a $K=3$ problem, etc.

\begin{figure}[thb]
    \centering
    \includegraphics[width=1\linewidth]{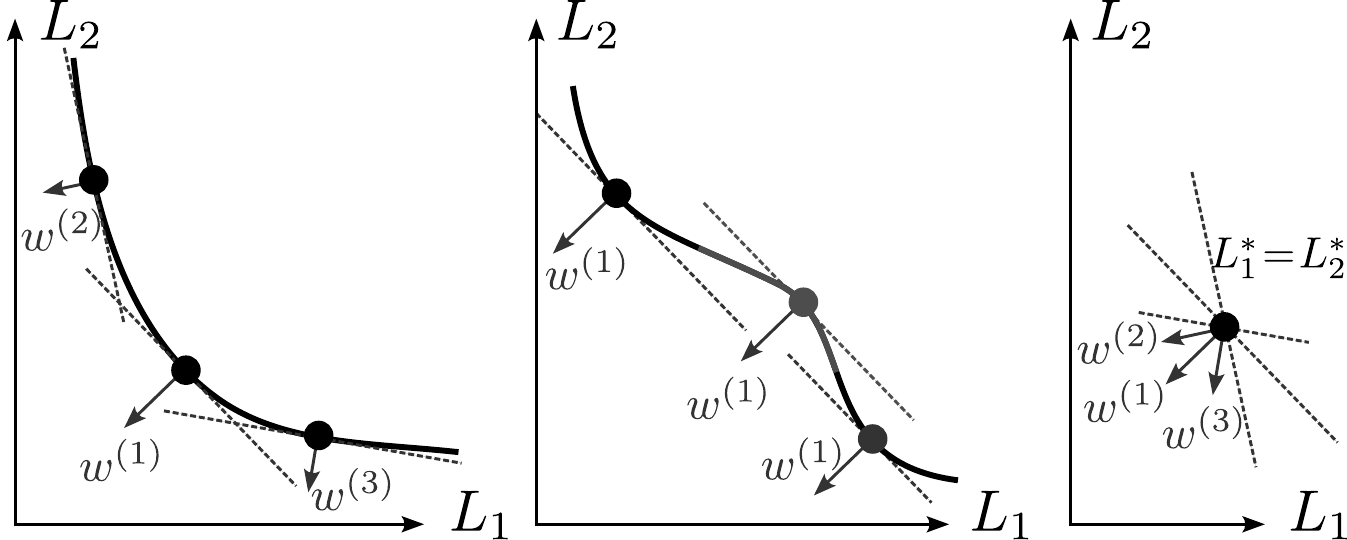}
    \caption{Left: Example of a convex Pareto front, where each point has a unique tangent, i.e., weighting vector $w^{(i)}$. Middle: Non-convex front, where multiple points have the same tangent vector. Right: Non-conflicting objectives such that the Pareto front collapses.}
    \label{fig:ParetoFronts}
\end{figure}

In deep learning, the most common situation is a very large number $q$ of trainable parameters, but a moderate number $K$ of objectives. It is thus much more common to visualize and study Pareto fronts instead of Pareto sets.
Figure \ref{fig:ParetoFronts} shows three examples of a convex and a non-convex problem, as well as the special case of non-conflicting criteria.

\subsubsection{Gradients and optimality conditions}
Closely related to single-objective optimization, there exist first order optimality conditions, referred to as the \emph{Karush-Kuhn-Tucker (KKT)} conditions \cite{Miettinen1998}. A point $\theta^*$ is said to be \emph{Pareto-critical} if there exists a convex combination of the individual gradients $\nabla L_k$ that is zero. More formally, we have
\begin{align}
    \sum_{k=1}^K \alpha^*_k \nabla L_k(\theta^*) =0, \qquad \sum_{k=1}^K \alpha^*_k =1, \label{eq:KKT} \tag{KKT}
\end{align}
which is a natural extension of the case $K=1$. 

Equation \eqref{eq:KKT} is the basis for most gradient-based methods (more details in Section \ref{para:IndividualParetoOptima}) and thus essential in the construction of efficient multi-objective deep learning algorithms. Their goal is to compute elements from the Pareto critical set,
\[
    \Pset_c = \condSet{\theta^*\in\R^q}{\exists \alpha^*\in\R_{\geq 0}^{K},\, \sum_{k=1}^K \alpha^*_k =1\,:~ \eqref{eq:KKT}\,\mbox{holds}},
\]
which contains excellent candidates for Pareto optima, since $\Pset_c \supseteq \Pset$. In a similar fashion to the single-objective case, there exist extensions to constraints \cite{GPD17,Fliege2016}, but we will exclusively consider unconstrained problems here.

\begin{remark}[Lipschitz continuous loss functions]\label{rem:Lipschitz}
    An extension is required if we consider less regularity, i.e., Lipschitz continuity. In the context of deep learning, this is the case for activation functions with kinks (e.g., $\operatorname{ReLU}(x)=\max\{0,x\}$), or when considering $\ell_1$ regularization terms ($\|\theta\|_1 = \sum | \theta_i |$), cf.\ Figure \ref{fig:ReLU_L1}. Then, \emph{subdifferentials} $\partial L_k$ take the place of the gradients $\nabla L_k$ (cf.\ \cite{Clarke1990} for a detailed introduction), and the condition \eqref{eq:KKT} is replaced by a non-smooth KKT condition
\begin{equation}\label{eq:KKT_Lipschitz}
    0\in \operatorname{conv} \left(\bigcup_{k=1}^K \partial L_k(\theta^*) \right),
\end{equation}
which reduces to \eqref{eq:KKT} for all $\theta^*$ where the $L_k(\theta^*)$ are smooth, i.e., outside $\theta=0$ in the exemplary Figure \ref{fig:ReLU_L1} below.
\begin{figure}[h!]
    \centering
    \includegraphics[width=.5\linewidth]{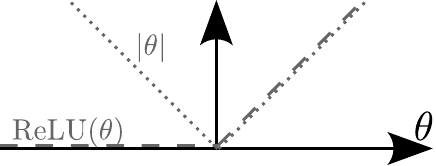}
    \caption{$\ell_1$ norm and ReLU activation function for $\theta\in\R$.}
    \label{fig:ReLU_L1}
\end{figure}
\end{remark}

\subsubsection{Overview of methods}
There exists a large number of conceptually very different methods to find (approximate) solutions of \eqref{eq:MOP}. The first key question is---and this will come up again in our taxonomy---whether we put the \emph{decision-making}\footnote{Decision-making refers to the selection of a particular element of $\Pset$, either once or interactively to react to changing circumstances. It is a topic of its own \cite{Triantaphyllou2000}, but we will not further study the decision-making process here.} before (\ref{para:IndividualParetoOptima}) or after (\ref{para:EntireSet}) optimization, or even consider an interactive approach (\ref{para:Interactive}).

\paragraph{Computing individual Pareto optima}
\label{para:IndividualParetoOptima}
When computing a single Pareto optimal (or critical) point, this means that some decision has been made beforehand to guide which point $\theta^*$ is sought. The two main approaches are to actively decide by means of a parameter-dependent \emph{scalarization}, or to accept any optimal point, i.e., to leave the decision-making to chance. 

In the former case, we synthesize the loss vector $L(\theta)$ into a single loss function $\hat{L}(\theta)\in\R$, meaning that we scalarize the optimization problem, see \cite{Miettinen1995,Ehr05} for detailed overviews. As a consequence, we can leverage the entire literature on single objective algorithms such as gradient descent or (Quasi-)Newton methods.

\begin{figure}[thb]
    \centering
    \includegraphics[width=.7\linewidth]{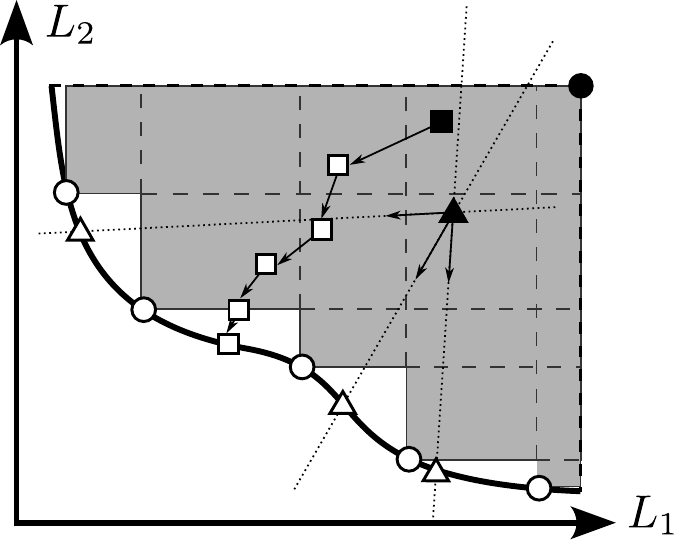}
    \caption{Sketch of different multiobjective optimization concepts. Entire front $\Pfront$ via hypervolume maximization (circles) or single point via gradient descent (squares) or preference vector scalarization (triangles).}
    \label{fig:MOMethods}
\end{figure}

Scalarization requires the selection of a set of weights $w$. In many cases, we have $w\in\R^K$, i.e., as many weights as we have objectives. The simplest technique is the \emph{weighted sum}
\begin{equation}
    \hat{L}(\theta) = \sum_{k=1}^K w_k L_k(\theta), \label{eq:WS}\tag{WS}
\end{equation}
which is also implicitly used in all sorts of regularization techniques (e.g., $\min_{\theta} L(\theta) + \lambda \|\theta\|_2^2$, $\lambda\in\R_{>0}$. Then, $w_1 = 1/(1+\lambda)$ and $w_2 = \lambda/(1+\lambda)$).
Geometrically, the weights $w$ define hyperplanes determining which point on the front we will find. This is visualized in Figure \ref{fig:ParetoFronts}, where we also see that non-convex problems immediately result in non-uniqueness. There are thus many, more sophisticated alternatives to \eqref{eq:WS} such as the $\epsilon$-constraint method, where all objectives but one are transformed into constraints:
\begin{equation}
    \min_{\theta\in\R^q} L_k(\theta)\quad \mbox{s.t.}\quad L_i (\theta)\leq \epsilon_i ~\forall~i \in \{1,\ldots,K\} \setminus k. \label{eq:epsConstraint}
\end{equation}
Alternatively, preference vectors \cite{Thiele2009,Eichfelder2009} (also known under the name \emph{Pascoletti-Serafini} \cite{PS84}) are highly popular. This approach is visualized by the triangles in Figure \ref{fig:MOMethods}, where the optimization problem is transformed into stepping as far as possible along the direction defined by the preference vector. While these advanced techniques beyond \eqref{eq:WS} are capable of handling non-convex problems, they usually introduce more challenging single-objective problems due to additional constraints (cf.\ \eqref{eq:epsConstraint}).

As an alternative, \emph{multiple-gradient descent algorithms (MGDAs)} are increasingly popular, in particular when it comes to very high-dimensional problems. The key ingredient is the calculation of a \emph{common descent direction} $d(\theta)\in\R^q$ that satisfies
\[
    \left(\nabla L_k(\theta)\right)^\top d(\theta) < 0, \quad k\in\{1,\ldots,K\},
\]
which again is a straightforward extension of single-objective descent directions. The determination of such a $d$ usually requires the solution of a subproblem in each step, for instance a quadratic problem of dimension $K$ \cite{Schaeffler2002,Dsidri2012},
\begin{equation}\tag{CDD}\label{eq:CDD}
\begin{aligned}
    d(\theta) &= -\sum_{k=1}^K w_k \nabla L_k(\theta),\qquad \mbox{where}  \\
    w&=\arg\min_{\begin{array}{c}\scriptstyle{\hat{w}\in[0,1]^K} \\ \scriptstyle{\sum \hat w_k=1} \end{array}} \left\|\sum_{k=1}^K \hat w_k \nabla L_k(\theta)\right\|_2^2.
\end{aligned}
\end{equation}
However, there are various alternatives to \eqref{eq:CDD} such as a dual formulation \cite{FS00} or a computationally cheaper so-called \emph{Franke-Wolfe} approach \cite{SK18}. Once a common descent direction $d(\theta)$ has been obtained, we proceed in a standard fashion by iteratively updating $\theta$ until convergence or some other stopping criterion is met, cf.\ Algorithm \ref{alg:MGDA} as well as the squares in Figure \ref{fig:MOMethods} for an illustration.
\begin{algorithm}[thb]
\caption{Multiple-gradient descent algorithm (MGDA)}\label{alg:MGDA}
\begin{algorithmic}[1]
\Require Initial guess $\theta^{(0)}$, learning rate $\eta \in\R_{>0}$ (possibly adaptive), maximum number of iterations $i_{\mathsf{max}}$, hyperparameters (depending on specific version of MGDA)
\Ensure $\theta^*\in\Pset_c$
\State Set $i=0$
\While{$\theta^{(i)}\notin\Pset_c$ \textbf{and} $i<i_{\mathsf{max}}$}
    \State Calculate gradients $\nabla L_i\left(\theta^{(i)}\right)$ for $i=1\ldots,k$
    \State Calculate descent direction $d\left(\theta^{(i)}\right)$ (e.g., via \eqref{eq:CDD})
    \State If adaptive, determine learning rate $\eta\left(\theta^{(i)}\right)$
    \State Update $\theta$:
    \[
        \theta^{(i+1)}= \theta^{(i)} + \eta\left(\theta^{(i)}\right) d\left(\theta^{(i)}\right)
    \]
    \State $i = i + 1$
\EndWhile
\end{algorithmic}
\end{algorithm}
Various extensions concern Newton \cite{Fliege2009} or Quasi-Newton \cite{Povalej2014} directions, uncertainties \cite{PD18b}, momentum \cite{Tanabe2023,SP24a,Nikbakhtsarvestani2023}, or non-smoothness \cite{Miettinen1995,GP21,Tanabe2018,Tanabe2023}. 

As laid out at the beginning of this section, this approach omits the decision-making. MGDAs yield Pareto critical points $\theta^*\in\Pset_c$, but one usually cannot determine which one, and how the different goals are prioritized in that point. To achieve this, one needs to resort to hybrid approaches including, e.g., preference vectors \cite{Zhang2024}.

\paragraph{Computing the entire set}
\label{para:EntireSet}
If we want to postpone the decision-making to take a more informed decision, we need to calculate the entire Pareto set $\Pset$ and front $\Pfront$. The most straightforward approach is to adapt the weights $w$ in scalarization and solve the single-objective problem multiple times. Alternatively, one can combine MGDA with a multi-start strategy (i.e., a set of random initial guesses $\left\{\theta^{(0,j)}\right\}_{j=1}^M$) to obtain multiple points. However, in both cases, it may be very hard or even impossible to obtain a good coverage of $\Pset$, i.e., that approximates the entire set with evenly distributed points.

\begin{remark}[Box coverings]
    An alternative to approximating $\Pset$ by a finite set of points is to introduce an outer box covering (see, e.g., \cite{Schtze2003,Dellnitz2005}). In theory, the numerical effort for a suitably fine covering grows exponentially with the dimension of the object we want to approximate (i.e., with the number of objectives $K$), but is independent of the parameter dimension $q$. This is good for the common case of few objectives. However, in practice, set-based numerics often also scale with $q$ due to the need to represent the boxes via Monte Carlo sampling, thus rendering them too expensive for applications in machine learning.
\end{remark}

Instead of parameter variation or using multi-start, we can directly consider a \emph{population} of weights $\left\{\theta^{(j)}\right\}_{j=1}^M$ that we iteratively update to improve each individual's performance while also ensuring a suitable spread over the entire front $\Pfront$. The population's performance is often measured by the hypervolume metric (see, e.g., \cite{Auger2012}), which is also visualized in Figure \ref{fig:MOMethods}. This is defined as the union of the boxes spanned by some reference point (shown in black) and one of the population's individuals, respectively.
One can attempt to directly maximize this metric, for instance using Newton's method \cite{Sosa2014}, see also the survey \cite{Bader2011} for for an extensive introduction.

The more popular alternative when optimizing an entire population is via \emph{multi-objective evolutionary algorithms (MOEAs)} \cite{CLV07}, see Figure \ref{fig:MOEA} for an illustration and Algorithm \ref{alg:MOEA} for a rough algorithmic outline. Therein, the pupulation's fitness is increased from generation to generation by maximizing a criterion that combines optimality (or non-dominance) with a spreading criterion. The most popular and widely used algorithm in this category is likely NSGA-II \cite{Deb2002}, but there are many alternatives regarding the crossover step (Step 3 in Algorithm \ref{alg:MOEA}), mutation (Step 4), the selection (Step 5), the population size, and so on. For more details, see the surveys \cite{Fonseca1995,Zhou2011,Tian2021}. Finally, many combinations exist with---for instance---preference vectors \cite{Thiele2009} or gradients \cite{Bosman2012,Nikbakhtsarvestani2023}, such that the creation of offspring is more directed using gradient information.
\begin{figure}[thb]
    \centering
    \includegraphics[width=.9\linewidth]{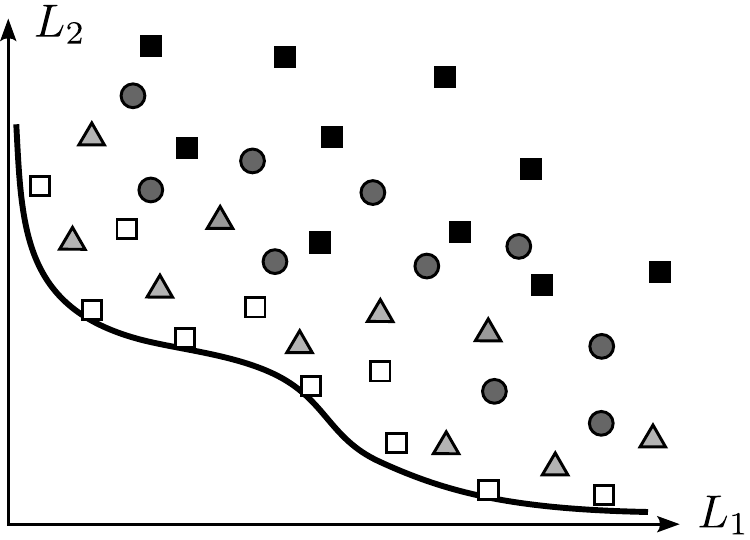}
    \caption{MOEA example, where a population of individuals is improved from one generation to the next ($\blacksquare\rightarrow\bigcirc\rightarrow\triangle\rightarrow\square$).}
    \label{fig:MOEA}
\end{figure}

\begin{algorithm}[thb]
\caption{Multi-objective evolutionary algorithm (MOEA)}\label{alg:MOEA}
\begin{algorithmic}[1]
\Require Initial population $P(0)$ of individuals $\left\{\theta^{(0,j)}\right\}_{j=1}^M$, number of generations $i_{\mathsf{max}}$,  hyperparameters (depending on specific version of MOEA)
\State Set $i=0$
\While{$i<i_{\mathsf{max}}$}
    \State Create an offspring population $\widehat{P}{(i)}$ out of ${P}{(i)}$, e.g., using \textbf{crossover} between two individuals
    \State Modify offspring population via \textbf{mutation}:
    \[\widetilde{P}{(i)} = \mathcal{M}\left( \widehat{P}{(i)} \right)\]
    \State \textbf{Selection} of the next generation ${P}{(i+1)}$ either from $\widetilde{P}{(i)}$ or from ${P}{(i)} \cup \widetilde{P}{(i)}$ (the latter is called \emph{elitism}) by a survival-of-the-fittest process (e.g., using a non-dominance and spread metric)
    \State $i = i + 1$
\EndWhile
\end{algorithmic}
\end{algorithm}

A final technique falling into the category of approximating the entire Pareto set by a finite set of points is \emph{continuation}. 
Rewriting the condition \eqref{eq:KKT} as a zero-finding problem,
\[
    H(\theta^*, \alpha^*) = \begin{pmatrix}
        \sum_{k=1}^K \alpha^*_k \nabla L_k(\theta^*) \\ \sum_{k=1}^K \alpha^*_k - 1
    \end{pmatrix} = 0,
\]
we find that under suitable regularity assumptions (i.e., twice continuously differentiable losses), the implicit function theorem says that the zero level set of $H$ is a smooth manifold of dimension $K-1$ \cite{Hil01}. The tangent space can be computed from the kernel of the Jacobian $H'(\theta^*, \alpha^*)\in\R^{q+1 \times q + K}$,
\[
    \resizebox{\linewidth}{!}{$H'(\theta^*, \alpha^*) = \begin{pmatrix}
        \sum_{k=1}^K \alpha^*_k \nabla^2 L_k(\theta^*) & L_1(\theta^*) & \hdots & L_K(\theta^*) \\ 0 \quad \hdots \quad 0 & 1 & \hdots & 1
    \end{pmatrix}.$}
\]
The procedure is then to start from a known Pareto optimum, compute a predictor within the tangent space and then compute the next Pareto critical point through a corrector step (e.g., using MGDA), see Figure \ref{fig:Continuation} and Algorithm \ref{alg:Continuation}.
However, in addition to the gradients, we also require the Hessians of all loss functions. 
These are $q \times q$ matrices and thus expensive to calculate as well as to store. In addition, the $C^2$ regularity is violated by many network architectures (e.g., when using ReLU activations) or loss functions.
An extension to Lipschitz continuous objectives can be found in \cite{BGP22}, even though it is computationally even more expensive.

\begin{figure}[thb]
    \centering
    \includegraphics[width=\linewidth]{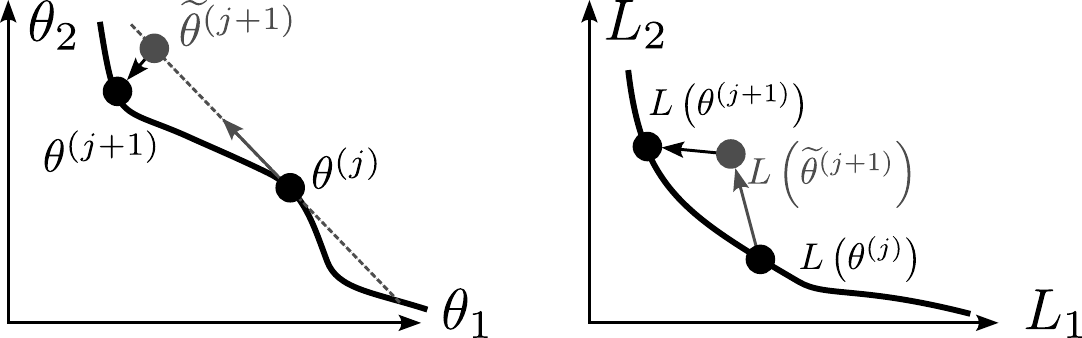}
    \caption{Continuation algorithms use predictor steps $\widetilde{\theta}$ along the tangent space of $\Pset_c$, i.e., in parameter space (left). A corrector step then produces the next Pareto optimal point. The right plot shows the corresponding points in the objective space.}
    \label{fig:Continuation}
\end{figure}

\begin{algorithm}[thb]
\caption{Continuation}\label{alg:Continuation}
\begin{algorithmic}[1]
\Require Initial Pareto critical point $\theta^{(0)}$ and KKT multiplier $\alpha^{(0)}$,  hyperparameters
\State Set $j=0$
\While{other end of front has not been reached}
    \State Compute tangent space in $\theta^{(j)}$ using kernel vectors of the weighted Hessian matrix $H'(\theta^{(j)},\alpha^{(j)})$
    \State \textbf{Predictor} step along the tangent space $\rightarrow$ $\widetilde\theta^{(j+1)}$
    \State \textbf{Corrector} step to obtain the next Pareto critical point $\theta^{(j+1)}$
    \State $j = j + 1$
\EndWhile
\end{algorithmic}
\end{algorithm}

\paragraph{Interactive methods}
\label{para:Interactive}
Regardless of the order of decision-making and optimization, all previously mentioned approaches can be implemented in a block-wise manner, i.e., a single run of an algorithm. Instead, \emph{interactive} methods (e.g., \cite{Miettinen1995,EMKH10,Kuefer2003,SCM+20}) alternate between decision-making and optimization. The approach usually starts from a Pareto optimum. Then, based on the preference of the decision maker (e.g., ``improve objective $L_1$, do not get worse in $L_2$, but a drop in $L_3$ is acceptable''), we compute another Pareto optimum that respects this prioritization. Very naturally, these algorithms have a close relation to continuation methods, where the decision maker's preference needs to be translated into a suitable predictor direction in the tangent space. 

\subsection{Multi-objective machine learning}\label{subsec:MOML}
The combination of multi-objective optimization and machine learning has been studied for several decades already, see \cite{Jin2006,JinSendhoff2008} for overviews. While the combination is a very natural one due to various performance criteria that are relevant in learning models from data, the focus has mostly been on other types of machine learning than deep neural networks. From the authors' point of view, the main reason is the large computational cost that comes with both multi-objective optimization and deep learning, which renders their combination very challenging. In Table \ref{tab:ProCon}, we list the main pros and cons of various MO techniques when it comes to deep learning, in particular concerning the large computational cost.

\begin{table}[htb]
    \centering
    \caption{Pros and cons of various classes of multi-objective optimization algorithms. The main computational challenges in the context of deep learning are shown in \textbf{bold}.}
    \label{tab:ProCon}
    \begin{tabularx}{\linewidth}{c|l|X}
        \textbf{Algorithm} & \hfil \textbf{Advantages}  & \hfil \textbf{Challenges} \\
        \textbf{class} & &  \\ \hline
        MOEAs & $\bm{+}$ often gradient-free & $\bm{-}$ \textbf{large \# of fcn.\ evals.} \\
        & $\bm{+}$ global optimization & $\bm{-}$ slow convergence \\ \hline
        Scalarization & $\bm{+}$ single-objective opt. & $\bm{-}$ parametrization via $w$ \\
        & & $\bm{-}$ \textbf{additional constraints} \\ \hline
        MGDA & $\bm{+}$ convergence similar & $\bm{-}$ no steering \\
        & to single-objective opt. & $\bm{-}$ \textbf{sub-routine for direction} \\
        & & $\bm{-}$ single optimum \\ \hline
        Continuation & $\bm{+}$ fast convergence & $\bm{-}$ req.\ smoothness: $C^2$ \\
        & & $\bm{-}$ \textbf{Hessian calculation} \\
        & & $\bm{-}$ only connected fronts
    \end{tabularx}
\end{table}

Thus---before turning our focus on deep learning in the next section---we here want to give a brief list of ways to incorporate multi-objective optimization with machine learning:
\begin{itemize}
    \item Feature selection \cite{Shu2016, gong2015, al2024multi},
    \item hyperparameter tuning \cite{Karl2023,MoralesHernndez2022},
    \item architecture search \cite{Jin.etal.2007},
    \item data imputation \cite{Lobato2015}
    \item training with respect to multiple objectives, for instance, support vector machines \cite{Suttorp2006,Akan2013}, decision trees, bayesian classifiers, radial basis function networks, or clustering (see also the references in \cite{Jin2006,JinSendhoff2008,Alexandropoulos2019}).
    \item multi-objective clustering \cite{alok2015new, gonzalez2020improving, mitra2020unified}
\end{itemize}
As in particular the last point sets deep learning apart from other machine learning techniques, this will be the focus of our taxonomy in the next Section.

\begin{remark}
    It should be noted that there already exist several surveys on the topic of multi-objective machine learning, specifically in the context of MOEAs \cite{Alexandropoulos2019,Tian2021} and hyperparameter tuning \cite{MoralesHernndez2022,Karl2023}.
    Moreover, three articles have introduced taxonomies of multi-objective reinforcement learning algorithms \cite{Liu2015,Hayes2022,Felten2024}, which share similarities with our classification in the next section. Nevertheless, we believe that this article closes a gap in particular in the areas of deep neural network training and gradient-based approaches (i.e., MGDAs), which are still less popular in the optimization community than MOEAs.
\end{remark}

\section{A taxonomy of multi-objective deep learning}\label{sec:Taxonomy}
Before we introduce our taxonomy for the multi-objective training procedure of deep neural networks in Section \ref{subsec:TaxonomyTraining}, let us first distinguish between the various purposes of multi-objective optimization in deep learning. These are, in no particular order,

\begin{enumerate}[label=(\roman*)]
    \item preprocessing steps such as feature selection or data imputation,
    \item the treatment of multiple primary tasks such
    \begin{itemize}
        \item multi-task learning,
        \item multi-class classification,
        \item clustering with respect to multiple criteria,
        \item multiple rewards in reinforcement learning,
    \end{itemize}
    \item the consideration of secondary tasks, e.g.,
    \begin{itemize}
        \item regularization,
        \item sparsity,
        \item fairness,
        \item interpretability,
        \item incorporation of prior knowledge,
    \end{itemize}
    \item using multi-objective optimization as a tool to improve the performance of a deep learning task.
\end{enumerate}
While these are quite versatile tasks, they share the common structure that a multi-objective optimization problem of the form \eqref{eq:MOP} needs to be solved, for which one can pursue various strategies.

\subsection{Taxonomy of the training procedure}
\label{subsec:TaxonomyTraining}
\begin{figure*}[thb]
    \centering
    \includegraphics[width=.7\linewidth]{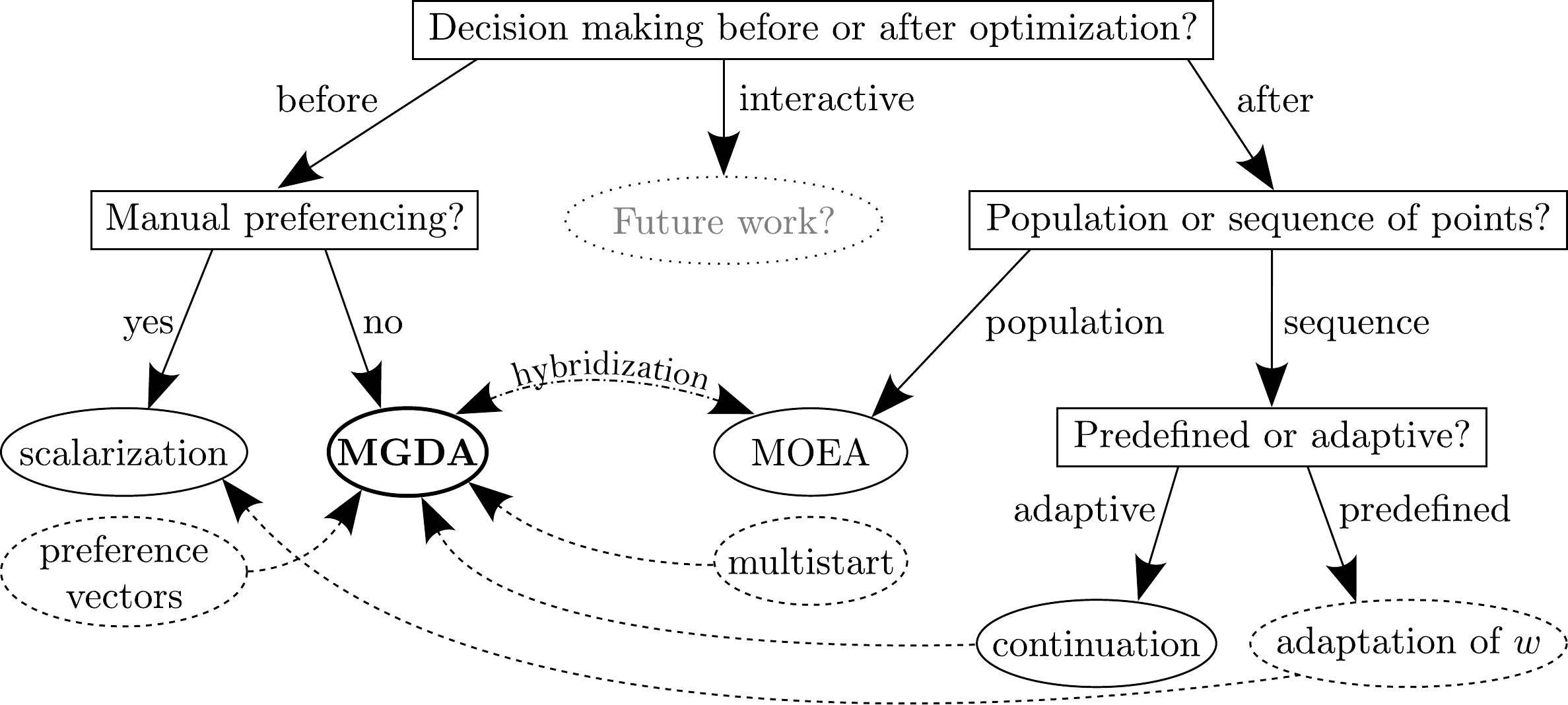}
    \caption{Taxonomy of methods. The squares denote design decisions, the ellipses denote classes of algorithms. Dashed ellipses refer to extensions building on other algorithms. 
    In the context of deep learning, gradient-descent (i.e., \textbf{MGDA}) lies at the heart of most successful approaches.}
    \label{fig:Taxonomy}
\end{figure*}

Figure \ref{fig:Taxonomy} shows our taxonomy of multi-objective optimization methods for deep learning problems. In the following, let's discuss various alternatives while taking into account the advantages and challenges highlighted in Table \ref{tab:ProCon}.

Before diving in, it should be noted that scalarization takes a special role since we essentially eliminate the multi-objective nature of the problem, which allows us to leverage many techniques from single-objective optimization. In particular, the weighted sum approach \eqref{eq:WS} introduces no additional constraints and is implicitly being used whenever penalty terms are added, e.g., for regularization or sparsity. $\epsilon$ constraint \eqref{eq:epsConstraint} or the Pascoletti-Serafini render the optimization significantly more challenging due to the constraints, which is why they are much less common in the context deep learning.

Turning our attention to the taxonomy in Figure \ref{fig:Taxonomy}, the first decision one has to take is whether the entire Pareto front is desired, or if we are instead satisfied with a single optimal compromise. The former results in substantial additional cost, which is why the deep learning community has until now mostly opted for the ``decide-then-optimize'' strategy. Besides the just-mentioned simple weighted-sum approach, MGDA is by far the most popular option, since---aside from the need to solve \eqref{eq:CDD} (or a similar sub-problem) instead of simply calculating a single gradient---it allows us to use all the existing machinery for gradient-based training, i.e., momentum, stochastic gradients, etc.

In the case of ``optimize-then-decide'', there currently appear to be two similarly popular approaches. First, evolutionary algorithms are extremely popular and at the same time easy to use, which is why they are a natural choice for a first attempt at multi-objective optimization. However, due to their large cost and otherwise slow convergence, a hybridization with gradients is advisable in the context of deep learning. Second, scalarization problems can easily be used to find multiple Pareto optima simply by varying the weight vector $w$. This has been used extensively, in particular in combination with the weighted sum. When it comes to continuation, there is until now little work, as the computational cost associated with second order information can quickly become prohibitively large.

The third option of interactive multi-objective deep learning has---to the best of our knowledge---not yet been explored with multiple objectives. Research in this area is also known as \emph{interactive machine learning (IML)} \cite{Bian2021,Budd2021} or \emph{human-in-the-loop (HITL)} \cite{Wu2022,MosqueiraRey2022}. However, most concepts are related to active learning or to human intermediate tasks such as data annotation, or human feedback for interpretability or performance improvement.\footnote{In particular in the literature on large language models, human feedback for fine tuning is an important contribution.}
A research direction such as interacting with a decision maker in terms of design criteria---as is popular in other multi-objective contexts \cite{Miettinen1995,Kuefer2003,EMKH10,SCM+20}---remains a task for future research.

\begin{remark}[Special role of reinforcement learning]\label{rem:TaxonomyRL}
    Due to its sequential decision-making character, reinforcement learning is fundamentally different from the other learning paradigms. The goal is to find a policy $\pi$ from interaction with a system, which means that presenting a Pareto front to a human decision maker is not the scenario RL is intended for. Common themes in multi-objective RL are thus decide-then-optimize (i.e., the left branch of Figure \ref{fig:Taxonomy}), known under the term single-policy algorithms. Besides, multiple-policy algorithms follow the right branch in spirit, but the ``population'' in that context is usually a set of value or $Q$ functions. The decision-making is then performed online, based on an algorithmic decision rule or user-defined weighting. Due to this reason, we will cover this case separately in Section \ref{sec:MORL} as a special case.
\end{remark}

\section{Multi-objective deep learning survey}\label{sec:Survey}
Following the taxonomy in Figure \ref{fig:Taxonomy}, the goal of this section is to provide an overview of the current state of the art, and we are going to separate the contributions according to the different learning paradigms introduced in Section \ref{subsec:DL}. Before we dive in, we would like to highlight the various types of objectives that can be pursued with multi-objective optimization. There are multiple reasons to treat deep learning as a multi-objective problem, despite the increased cost:
\begin{enumerate}[label=(\roman*)]
    \item the consideration of multiple, equally important criteria, for example in the context of 
    \begin{itemize}
        \item multi-task learning, 
        \item multi-class classification,
    \end{itemize}
    \item trade-offs between different performance indicators, e.g., fairness versus bias, 
    \item balancing of knowledge and data, for instance in the field of physics-informed machine learning,
    \item inclusion of secondary objectives such as
    \begin{itemize}
        \item sparsity / regularization paths, 
        \item interpretability, 
    \end{itemize}
    \item multi-objective optimization as a performance-enhancing paradigm.
\end{enumerate}

\begin{remark}[A note on overparametrization]\label{rem:overparametrization}
    It is well known that many deep learning architectures possess such a large number of degrees of freedom that they can essentially fit any label structure. This phenomenon is known as \emph{overparametrization} \cite{Neyshabur2018}. A side-effect of this phenomenon in the context of multi-objective learning is that in the case of very powerful function approximators (i.e., large networks), one can in principle resolve the conflict between different objectives. This means that for some tasks such as multi-task learning, the Pareto front collapses to a single point, cf.\ Figure \ref{fig:ParetoFronts} on the right. An example of this can be found in the MultiMNIST example in \cite{SK18}.
\end{remark}

\subsection{Supervised learning}\label{subsec:MODL}

\subsubsection{Multiobjective Gradient Descent Algorithms}
The huge success of machine learning has also had a strong impact on the optimization community and the research directions pursued therein. For example, stochastic gradient descent has been widely studied. Even though these studies do not all consider deep learning problems, they have expensive problems (such as deep neural network training) in mind, for instance when developing multi-objective extensions of stochastic gradient descent \cite{Liu2021}, momentum-based algorithms for accelerated convergence \cite{Tanabe2023,SP24a,SP24b,Nikbakhtsarvestani2023}, or combinations thereof as in the multi-objective Adam algorithm \cite{Mitrevski2020}.

Other works directly consider deep learning applications, a very prominent example being \cite{SK18}. Therein, a so-called \emph{Frank Wolfe} routine was presented that replaces the sub-routine \eqref{eq:CDD} in Algorithm \ref{alg:MGDA} with a cheaper version. This method is then used to train a deep neural network with a shared parameter section for representation learning, followed by task-specific layers for the individual tasks (here the identification of multiple handwritten digits). Interestingly, there appears to be no conflict between the two tasks, which we believe is due to the overparametrization effect mentioned in Remark \ref{rem:overparametrization}.
Alternative gradient-based frameworks for deep multi-task are \emph{PCGrad} \cite{YTSASK20}---where each task's gradient is projected onto the normal plane of the other gradients to avoid expensive sub-problems such as \eqref{eq:CDD}--- or \emph{CAGrad} \cite{LBJSLQ21}, where \eqref{eq:CDD} is replaced by a (rather similar) formulation that tries to reduce conflicts between the individual tasks. In \cite{Liu2021b}, the \emph{Stein Variational Gradient Descent} (SVGD) algorithm from \cite{Liu2016} is extended to multiple objectives and tested on a large number of different problems such as accuracy-vs-fairness or multi-task learning. In the \emph{NashMTL} procedure \cite{NSAMK22}, multi-task learning is considered from a game-theoretic perspective, more precisely as a bargaining game.

\begin{figure}[htb]
    \centering
    \includegraphics[width=.5\linewidth]{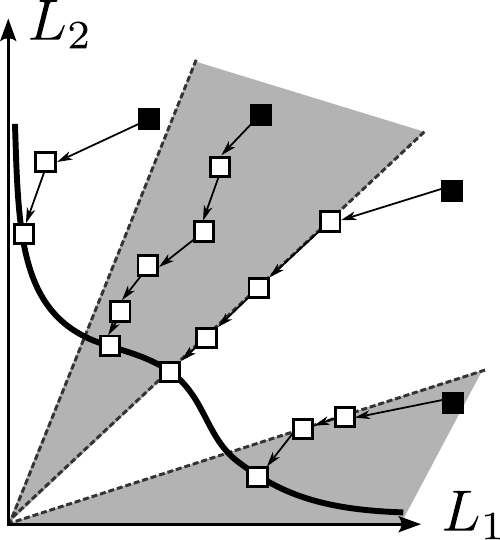}
    \caption{Pareto multi-task learning, where a population of points is optimized via MGDA, but each individual is restricted to an individual cone \cite{Lin2019}.}
    \label{fig:ParetoMTL}
\end{figure}

Several authors have extended the basic MGDA procedure to obtain an approximation of the entire front. For instance in \cite{Lin2019}, a multi-start, constrained version of MGDA is developed, where disjoint feasible cones ensure a good spread, see Figure \ref{fig:ParetoMTL} for an illustration. In \cite{Liu2021b}, Langevin dynamics are used to ensure a good spread within a population of individuals that are iterated via the SVGD algorithm.

\subsubsection{Scalarization}
In the area of scalarization, the vast majority of deep learning algorithms simply uses the weighted sum whenever more than a single loss is considered, such as additional regularization terms, or different loss contributions such as physics and data loss in physics-informed machine learning \cite{karniadakis2021physics}. Very often, this is simply done without naming it multi-objective optimization, as it is such a common theme. In the following, we will discuss scalarization techniques going beyond this weighted sum approach.
For instance, in \cite{LFSQ24}, the weighted sum is used but modified in terms of a dynamic weighting strategy over the iterations, such task losses are balanced efficiently. 
A so-called conic scalarizaiton techinque \cite{Kasimbeyli2011} which---loosely speaking---is a more sophisticated version of the weighted sum, is used in \cite{HP24} for multi-objective encoder training to balance performance and adversarial robustness in image classification.

To obtain a good coverage of the entire front, the weighted sum weights are adaptively selected via the so-called \emph{Non-Inferior Set Estimation} procedure in \cite{RDMLR21}, which is a type of bisection method so that the points are better distributed. As an example, they consider a multinomial loss versus $L_2$ regularization.
Weighted Chebyshev scalarization is used for multi-task learning in \cite{HBP24}, combined with proximal gradient descent to take sparsity into account as the third objective. 
In \cite{Ruchte2021} a weighted sum loss is balanced with a cosine similarity in order to improve the spread of points when varying the weight $w$.

\subsubsection{MOEAs}
As laid out earlier, MOEAs are in most cases too expensive for deep neural network training. The number of generations ($S$ in Algorithm \ref{alg:MOEA}) is usually very large, and so is the population size $M$. As a consequence, MOEAs tend to require a very large number of function evaluations. Moreover, the crossover procedure often consists of randomly combining two individuals, which is a close-to-hopeless procedure for the parameter dimensions we find in realistic deep leraning applications. As a consequence, the usage of MOEAs is restricted to hybridized versions (cf.\ our taxonomy in Figure \ref{fig:Taxonomy}) as is for instance done in \cite{Liu2021b}, where Langevin dynamics ensure a good spread of a population that is iterated using MGDA. Alternatively MOEAs can be used for other tasks in the context of deep learning such as neural architecture search (\cite{Elsken2019}, more details in Section \ref{subsec:NAS}) or the the selection between different networks. This was done in \cite{BDMPL22} for multiple  association rule learning to foster interpretability.

\subsubsection{Continuation}
Similar to MOEAs---but due to a different reason---continuation methods suffer from large computational cost that tends to be prohibitive for deep learning applications. 
We have seen in Section \ref{para:EntireSet} that continuation requires the Hessians of all loss functions, which are expensive to calculate as well as to store. 
To this end, research has mostly focused on very specific settings such as the regularization path between a main loss and the sparsifying $\ell_1$ norm $\|\theta\|_1$ \cite{Fu2020,BGP22,Fu2023}. Besides, there are two approaches trying to avoid Hessian computation. In \cite{Ma2020}, Hessians are approximated using the Krylov subspace method MINRES, whereas in \cite{APS23}, the minimization of one of the objectives is used as a suboptimal proxy for the predictor step.

\subsection{Unsupervised and self-supervised learning}\label{subsec:DUL}

When acquiring labeled data is expensive or impractical, unsupervised and self-supervised learning methods offer novel opportunities in deep learning to balance diverse objectives.  
\subsubsection{Scalarization} A foundational approach is the simple summation of losses, where individual losses are combined using fixed weights \cite{gui2019constructing, song2019metric, li2023multi, lu2024self}. 
Beyond straightforward loss aggregation, scalarization methods provide a more sophisticated way to balance objectives by converting them into a single scalarized loss. For instance, \cite{wan2023unsupervised} introduced an unsupervised active learning approach that integrates representativeness, informativeness, and diversity, while \cite{chen2021pareto} employed a Pareto self-supervised training approach to harmonize self-supervised and supervised tasks in few-shot learning. These methods highlight the potential of scaling objectives to address problem-specific priorities.

\subsubsection{Adaptive weighting}
Unlike scalarization, adaptive weighting dynamically adjusts the importance of objectives throughout the training process. 
For example, \cite{miranda2015multi} proposed self-adjusting weighted gradients to optimize the hypervolume in multi-task settings, demonstrating its application with denoising autoencoders. In \cite{ruan2022weighted}, it is explored how ensemble methods can enhance self-supervised learning by adaptively weighting losses across ensembled projection heads during training.

\subsubsection{Evolutionary algorithms} Since MOEAs tend to be too expensive for direct network optimization, \cite{liu2017structure} instead developed a structure-learning algorithm for deep neural networks based on multi-objective optimization, employing evolutionary strategies to identify optimal network architectures (see also the related Section \ref{subsec:NAS} on neural architecture search). Expanding on this, \cite{he2019evolutionary} integrated generative adversarial networks (GANs) with evolutionary multi-objective optimization, demonstrating how GANs can enhance the exploration of diverse solutions in high-dimensional spaces. Combining self-supervised representation learning with evolutionary search, \cite{xia2022molecule} developed a technique to discover optimal properties in an implicit chemical space.

Besides the above-mentioned references, the following studies address multi-objective optimization for unsupervised learning tasks such as feature selection \cite{gong2015, al2024multi, mierswa2006information, suman2019building, handl2006} and clustering \cite{gonzalez2020improving}, but they do not explore their applicability to deep neural networks.

\subsection{Generative modeling}\label{subsec:GenAI}
Generative modeling is currently one of the most popular areas of machine learning. Until now, the number of contributions with an explicit multi-objective treatment is rather limited, most of them in the area of generative adversarial networks. The works \cite{Durugkar2017,Albuquerque2019} consider multiple discriminator networks, even though they are not trained using multi-objective optimization. In \cite{Wang2022}, several GANs are weighted using the weighted sum method to have better control over the generated output. Adaptive weighted-sum-like approaches for multi-task training of multi-adversarial networks were developed in \cite{Han2023,Fu2024}, and multi-adversarial domain adaptation via the weighted sum in \cite{Pei2018}.

There are even fewer articles in other domains of generative modeling. In \cite{Wang2022}, a reference-point scalarization---very similar to reference vector methods---was used for multi-objective training of variational autoencoders that create data according to multiple criteria \cite{Wang2022}. Diffusion models were trained with respect to multiple criteria using MGDA in \cite{Yao2024}, and the multi-objective treatment of multiple judges for LLM feedback during the training phase was studied \cite{Xu2024}, where again weighted sum scalarization was used. It can thus be concluded that there is plenty of room for research in multi-objective generative modeling, be it with respect to more advanced optimization techniques beyond the weighted sum, or regarding advanced modeling approaches.

\subsection{Neural architecture search}\label{subsec:NAS}
Besides the direct use of multi-objective optimization algorithms for deep learning, which may be prohibitively expensive in many cases, one can also use multi-objective optimization on a meta level. Similar to algorithm selection techniques, \emph{neural architecture search} (see \cite{Elsken2019} for a detailed overview in the single-objective case) describes the task to select the best neural network architecture for a given learning problem with respect to criteria such as predictive performance, inference time, or number of parameters. In this area, MOEAs such as NSGA-II are a popular choice, e.g., \cite{Jin.etal.2007,Elsken2018,XZZWG24}. However, when using hypernetworks instead, this approach can be accelerated as well, for instance using MGDA \cite{SZSDGH24,KKHO21}.

\subsection{Applications}\label{subsec:Applications}
Before concluding, we here list a couple of applications, where multi-objective deep learning has proven to be helpful or even superior to the more established single-objective counterpart. This list is likely not exhaustive, but we will make an attempt to demonstrate the versatility of multi-objetive deep learning in various areas.

\subsubsection{Language and video analysis and enhancement}
Machine learning in the area of language, audio and video data is characterized by large amounts of data and long time series.
Applications of multi-objective concepts to speech include the multi-target training of long-short-term memory networks for speech enhancement \cite{Sun2017,Xu2017} or multi-objective speech recognition \cite{saif2024m2}.
In the context of video data, the summarization, streaming optimization and short video generation was studied \cite{dhanushree2024,ozccelebi2006} and \cite{xu2023short}, respectively.
For natural language processing, applications include text summation \cite{ryuetal2024}, text generation \cite{pour2024gaussian} prompt engineering for LLMs \cite{baumann2024}, meta-learning in terms of multi-objective LLM selection \cite{li2024s}, and the multi-objective treatment of multiple judges for LLM feedback during the training phase \cite{Xu2024}.
Finally, multi-objective adversarial gesture generation by weighting various discriminators was studied in \cite{Ferstl2019}.

\subsubsection{Engineering applications}
This section covers all sorts of technical applications beyond video and language. For instance multi-task learning was used for for phoneme detection in \cite{Seltzer2013}, a generative model for air quality and weather prediction was trained in \cite{Han2023}, medical image denoising by GANs trained with multiple discriminators was realized in \cite{Fu2024}, and multi-objective deep reinforcement learning was successfully used for workflow scheduling in \cite{Wang2019}.

\subsubsection{Physics-informed machine learning}
Finally, we would like to highlight the area of scientific machine learning, which has received tremendous attention in recent years. More specifically, \emph{physics-informed neural networks (PINNs)} \cite{karniadakis2021physics} are models that predict the solution of a differential equation. These are equations describing the dynamics of complex systems such as robots, fluid mechanics or nuclear fusion. The solution to such a differential equation is a function, the state of the system $u$ depending on time $t$ (and often on space $s$ as well), $u(t,s)$. The goal of PINNs is now to approximate $u(t,s)$ by a neural network that takes $t$ and $s$ as inputs, and produces $u$ as the output. Since the underlying equations are very often known, on can define a \emph{physics loss}, meaning that the output satisfies the differential equation at a number of random points in the space-time domain. Generalization then ensures that $f_\theta(t,s) \approx u(t,s)$ for all $t$ and $s$.

PINNs are a natural playground for multi-objective optimization, since we often have both physical knowledge and data, such that we have two losses that we would like to satisfy at the same time \cite{karniadakis2021physics}. For clean data and exact dynamics, these objectives are not in conflict \cite{AFSP24}, but this is seldom the case for real applications, where data is noisy, the system equations are approximations, or the domain of interest cannot be defined exactly (e.g., the flow inside the human heart). In this area, multi-objective optimization has the potential to become quite important, even though research has now been limited to weighted sum training \cite{Rohrhofer2023}, as well as a comparison between MGDA and MOEA \cite{AFSP24} for a few relatively small sample problems.

\section{Deep multi-objective reinforcement learning}\label{sec:MORL}

Due to its sequential decision-making nature, reinforcement learning (RL---see \cite{sutton2018reinforcement,bertsekas2019reinforcement} for excellent overviews) takes a special role in the field of machine learning, with concepts that often differ significantly from the other learning paradigms. Nevertheless, it also shares features such as heavy usage of deep neural network function approximators or gradient-based learning. In the following, we thus briefly cover the basics of RL, before addressing modifications to our taxonomy in the context of RL and surveying the literature on deep multi-objective RL.

\subsection{Basics}
\label{subsubsec:RL}
In contrast to supervised learning, RL follows a trial-and-error philosophy. That is, an agent interacts with its environment through actions $a\in\Acal$ and receives a reward $r\in\R$, indicating whether the action was beneficial or not. Through the action, the system state $s\in\Scal$ changes from time $t$ to $t+1$ in a probabilistic manner according to the transition operator $\Tcal: \Scal \times \Acal \rightarrow \mathcal{P}(\Scal)$. Since the dynamics is independent of past states\footnote{For instance, the transition in chess from one board position $s_t$ to the next position $s_{t+1}$ is independent of how $s_t$ was reached (i.e., of $s_{t-1}$, $s_{t-2}$, $\ldots$).}, this setting is referred to as a \emph{Markov Decision Process (MDP)}.

The goal in RL is to find a \emph{policy} $\pi: \Scal \rightarrow \mathcal{P}(\Acal)$ that maximizes the sum of discounted future rewards (with discount factor $\gamma\in(0,1]$), also referred to as the \emph{value}:
\begin{equation}
    V_{\pi}(s) = \E_{\pi}\left[\,\sum_{\tau=0}^{\infty}\gamma^{\tau}r_{t + \tau} \biggm\vert s_{t} = s \,\right]. \label{eq:Value-Function}
\end{equation}
A closely related concept is the so-called $Q$-function that determines the value of a state $s$ if we take action $a$ and then follow policy $\pi$,
\begin{equation}
    Q_{\pi}\left(s, a\right) = \E_{\pi} \left[ \, \sum_{\tau=0}^{\infty}\gamma^\tau r_{t + \tau}  \biggm\vert s_{t}=s, a_{t} = a  \, \right]. \label{eq:Q-Function}
\end{equation}
We thus have the relation $V_\pi(s) = Q_\pi(s, \pi(s))$.
Once we know $Q$, we can determine the optimal action by evaluating $Q$ for all $a$:
\begin{equation}\label{eq:Qselection}
    a^* = \arg\max_{a\in\Acal} Q_{\pi}\left(s, a\right).
\end{equation}
If the action set is not finite, but continuous, then \eqref{eq:Qselection} becomes a (potentially expensive) nonlinear optimization problem itself. In such a situation, we can instead try to learn the policy $\pi$ directly. Algorithms of this class usually make use of the \emph{policy gradient theorem}, where we directly compute gradients of a performance criterion (e.g., the integral over the value function) with respect to the policy, often using a so-called \emph{critic} to efficiently evaluate this gradient, see \cite{sutton2018reinforcement} for details.

To summarize, RL ultimately boils down to learning $V_\pi$, $Q_\pi$ or $\pi$ itself from experience, i.e., interactions with the environment. 
We are thus facing a dynamic optimization problem, where we iteratively update our behavior $\pi$ in order to maximize our value.
While this can conceptually be realized using the theory of Dynamic Programming \cite{bertsekas2019reinforcement} the complexity quickly supersedes all computing capacities. To circumvent this issue, deep reinforcement learning introduces neural network approximations of the above-mentioned functions. For discrete action spaces, \emph{deep $Q$-learning} (e.g., \cite{VanHasselt2016}) has proven very successful, whereas for continuous control tasks, policy gradient methods (e.g., the \emph{Proximal Policy Optimization (PPO)} \cite{schulman2017proximal}, the \emph{deep deterministic policy gradient (DDPG)} \cite{lillicrap2015continuous} or the \emph{Soft Actor Critic (SAC)} \cite{haarnoja2018soft}) are the most prominent methods.
There are numerous success stories of deep RL such as board or video games (Chess or Go \cite{Silver2018}, Atari \cite{Mnih2015}), robotics \cite{Kober2013} or complex physics systems \cite{Degrave2022,vignon2023recent,PSC+24}. In particular for continuous action spaces, these algorithms heavily rely on gradient descent, similar to the supervised learning case.

\subsection{MORL: adapted taxonomy and survey}\label{subsec:MORL}
\begin{figure}[htb]
    \centering
    \includegraphics[width=.6\linewidth]{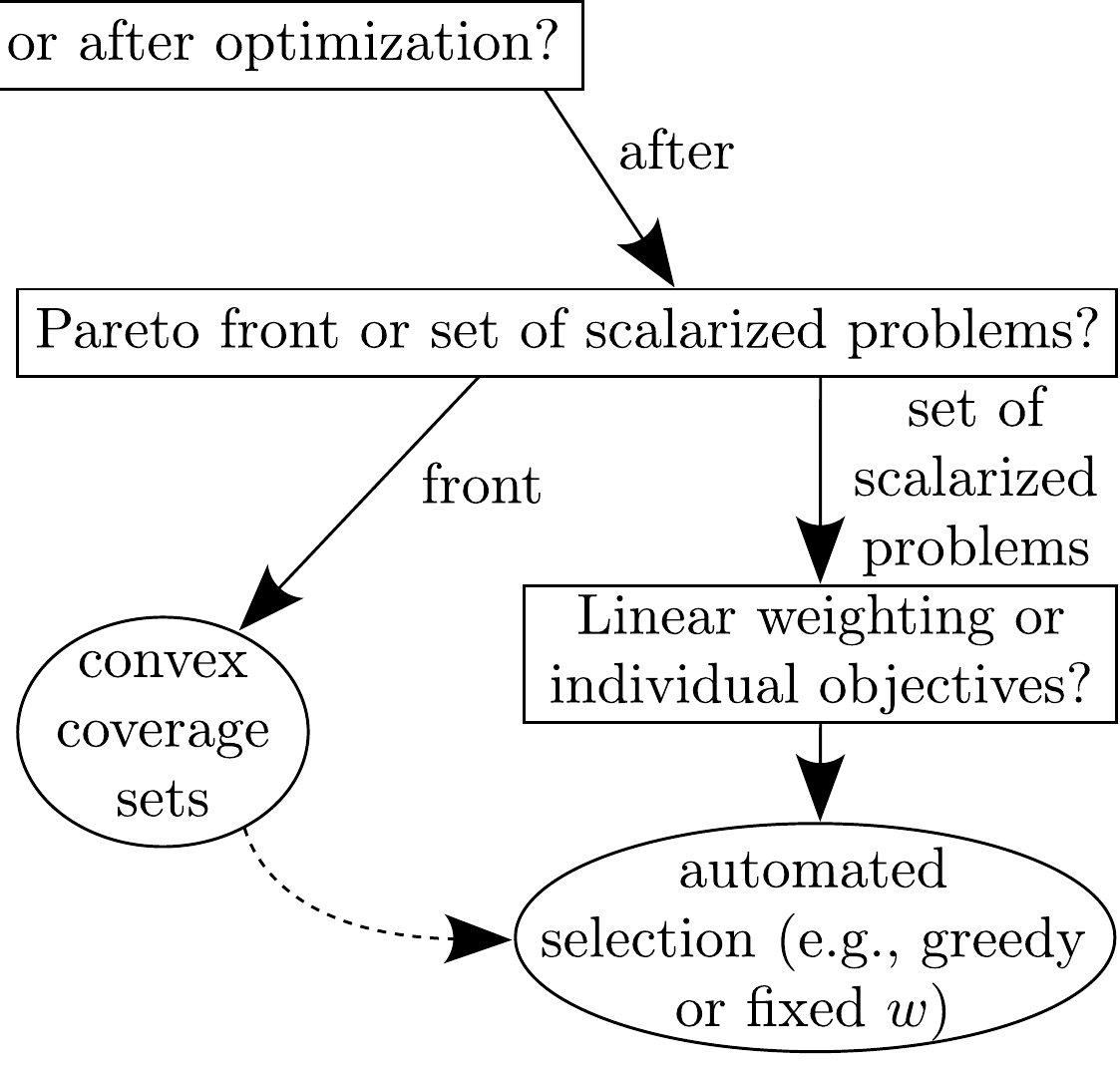}
    \caption{Modification of the right branch of Figure \ref{fig:Taxonomy} for MORL. 
    }
    \label{fig:TaxonomyRL}
\end{figure}

As mentioned in Remark \ref{rem:TaxonomyRL}, the taxonomy can only be partially applied to multi-objective reinforcement learning (MORL) due to its sequential decision-making nature.
In the MORL literature, the distinction between first-decide-then-optimize and first-optimize-then-decide---the top decision in our taxonomy in Figure \ref{fig:Taxonomy}---is also referred to as \emph{single-policy} versus \emph{multiple-policy} algorithms.
The central deviation from our taxonomy lies in the second option, where we do not present a Pareto set to a decision maker, but calculate multiple utility functions, which are then synthesized in an automated fashion to yield a Pareto optimal policy, cf.\ Figure \ref{fig:TaxonomyRL}.

Independent of which approach we pursue, the multi-objective setting means that we have a vector-valued reward $r\in\R^K$, as well as a vector-valued $Q$ function
\begin{equation}\label{eq:MO-Q-Function}
    Q_{\pi}(s,a) = \begin{pmatrix}
        Q_{\pi,1}(s,a) \\ \vdots \\ Q_{\pi,K}(s,a)
    \end{pmatrix},
\end{equation}
where the entries in \eqref{eq:MO-Q-Function} are as in \eqref{eq:Q-Function} but with the individual rewards $r_k$, respectively. The value function \eqref{eq:Value-Function} is transformed accordingly. 
Quite naturally, as the goal is always to increase the value, one can now define a Pareto set $\Pset$ of non-dominated value functions, see \cite{Roijers2013} for a detailed introduction.
It should be mentioned, though, that the multi-objective treatment introduces significant additional challenges in terms of finding Pareto optimal policies, see \cite{LHDY23} for details.

As mentioned earlier, there already exist several MORL overviews \cite{Vamplew2011,Roijers2013,Liu2015,Hayes2022}, which is why we are going to restrict our attention to the deep learning approaches. A general framework for the usage of various such deep MORL approaches has been presented in \cite{Nguyen2020}.
However, it must be said that conceptually, there is no real difference between deep MORL and other MORL algorithms---the conceptual treatment of multiple criteria is entirely in the synthesization of the policy, and thus largely independent of how the individual value or $Q$ functions are modeled/approximated. 

The largest part of the MORL literature is concerned with $Q$ learning, i.e., learning approximations of $Q$ functions of the form \eqref{eq:MO-Q-Function} and then selecting the action by solving a multi-objective extension of \eqref{eq:Qselection}.
A large number of techniques relies on scalarizing the vector-valued $Q$ function \eqref{eq:MO-Q-Function} in a linear fashion using a weight vector $w\in\R^K$,
\begin{align}
    \hat{Q}_{\pi}(s,a) &= w^\top  Q_{\pi}(s,a) = \sum_{k=1}^K w_k Q_{\pi,k}(s,a) \notag \\
    &= \sum_{k=1}^K\E_{\pi}\left[\, \sum_{\tau=0}^{\infty}\gamma^{\tau}r_{k,t + \tau} ~\middle\vert~ s_{t} = s, a_{t} = a \,\right] \label{eq:ScalarizedQ} \\
    &=\E_{\pi}\left[\, \sum_{\tau=0}^{\infty}\gamma^{\tau} \left( \sum_{k=1}^K r_{k,t + \tau} \right) ~\middle\vert~ s_{t} = s, a_{t} = a \,\right]. \notag
\end{align}
The distinction of single-policy versus multiple-policy then boils down to the question of when and how to select $w$. Options are:
\begin{enumerate}[label=(\roman*)]
    \item \textbf{single-policy}: fix $w$ ahead of time (nonlinear versions of scalarization equally possible)
    \item \textbf{single-policy}: determine a rule according to which $w$ is adjusted dynamically
    \item \textbf{multiple-policy}: train multiple $\hat Q^{(j)}$ (and consequently, $\hat V^{(j)}$) corresponding to weights $\{w^{(j)}\}_{j=1}^M$. This can also refer to individual objectives (i.e., $w=[1,0,\ldots,0]$). Optionally, one can then form the so-called \emph{convex coverage set} (CCS) \cite{Roijers2013,Roijers2015} that interpolates linearly between the non-dominated value functions.\footnote{Conceptually, there is a close connection to explicit model predictive control \cite{Bemporad2002}, where we also compute a simplex of feedback laws based on a finite set of control problems.} For weighting, we have several options:
    \begin{enumerate}[label=\alph*),itemindent=*]
        \item fixed weights during planning
        \item dynamically adapted weights during planning
        \item selection policy (e.g., greedy)
    \end{enumerate}
\end{enumerate}

\subsubsection{Single-policy algorithms}
In the single-policy setting with fixed weights (point (i) in the list above), any deep $Q$ learning architecture can readily be applied. In a similar fashion, the $\epsilon$-constraint method \eqref{eq:epsConstraint} can be transferred to the RL setting, where it is referred to as \emph{thresholded lexicographic ordering} (TLO). Since a constraint violation is undesirable, the constraint-objectives are considered more important, thus the ordering.
An alternative to avoid the decision-making is to allow non-linear scalarization and use, e.g., a neural network architecture to learn a suitable scalarizer $f_\theta(V(s)) = \hat{V}(s)$. Under the assumption that $f_\theta$ is strictly concave, one can show that the solution to this MDP is Pareto optimal \cite{Agarwal2022}.

For improved performance, a single-policy algorithm is proposed in \cite{van2014multi}, in which multiple single-objective $Q$-learning problems are solved for various Chebyshev scalarizations. During execution, a simple greedy strategy then selects the $Q$ function with maximal value, thus eliminating the decision-making.

Aside from scalarization, MGDA-like approaches have been proposed, for instance in \cite{KHPOS24,ZLKD22}, where---in the context of constrained reinforcement learning---policy gradients are aggregated into a single descent direction. In \cite{YLCHZ24}, multiple constraints are transformed into objectives and then considered using an MGDA-like procedure.

\subsubsection{Multiple-policy algorithms}
If we decide not to scalarize the reward or value or $Q$-function before training, then we can follow one of two strategies. The first one is to simply compute the vector of $V$ or $Q$ values and then decide online which one to take. This is realized in a linear fashion using a weight vector $w$ in, e.g., \cite{Abels2019}. In an extension of this work \cite{LHDY23}, the authors add a strictly concave term to the rewards, which overcomes issues with solutions that are not Pareto optimal.

A simple way to automatically choose a weight (in the context of $Q$ learning) is the so-called top-$Q$ approach \cite{Liu2015} where we always select the entry of $Q$ that maximizes the value, i.e.,
\[
    \max_{j} \max_{a\in\Acal} \hat{Q}^{(j)}(s,a).
\]
Again, we can use any deep RL approach to learn the individual $Q$ functions. This is in line with point (iii-b) from the enumeration above. A very similar strategy was followed in \cite{Mossalam2016}, where several scalarized problems of the form \eqref{eq:ScalarizedQ} are solved. Their values vary linearly with the weighting of the objectives such that during planning, we select the $Q$ network that has the largest value for the given $w$, to decide on the action. 

Again in a similar fashion to \cite{Agarwal2022}, the paper \cite{Tajmajer2018} considers a set of $Q$ functions, but at the same time learns a so-called decision value (using temporal difference learning) according to which the individual $Q$ values are weighted.

The multiple-policy section of \cite{van2014multi} treats the problem slightly differently, in that an entire set of non-dominated $Q$ functions is stored (i.e., those having non-dominated value for at least one action $a$). The selection of a strategy is then performed online, either following a decision makers preference or a greedy policy.

In \cite{WTKA22} several---seemingly unrelated---policies are trained using PPO, each taking into account different constraints. Depending on the decision maker's preference in terms of constraints, a suitable policy is then selected.

\section{Deep learning for multi-objective optimization}\label{sec:DLforMO}
Even though the content of this section is not at the center of our overview, we would like to briefly highlight the vice-versa combination of multi-objective optimization and deep learning, since the two are sometimes confused. Instead of treating deep learning problems using multi-objective optimization, one may also using deep learning to accelerate the solution of (classical) multi-objective optimization problems, such as the multi-criteria design of a complex technical system like an electric vehicle.

Very naturally, as solving MOPs is computationally expensive---in particular for complex and costly-to-evaluate models---there is a strong interest in accelerating the evaluation of the loss function $L(\theta)$ or their gradients.
There has been extensive research on surrogate-assisted multi-objective optimization \cite{Chugh2015,Tabatabaei2015,PD18a,Deb2021}. That is, instead of $L(\theta)$, we train a surrogate function $f_\phi(\theta)$---parametrized by $\phi$---using a small number $N_{\mathsf{exp}}$ of expensive model evaluations at carefully chosen points $\{\theta^{(i)}\}_{i=1}^{N_{\mathsf{exp}}}$:
\begin{equation}\label{eq:Surrogate}
    \min_\phi \sum_{i=1}^{N_{\mathsf{exp}}} \norm{L\left(\theta^{(i)}\right) - f_\phi\left(\theta^{(i)}\right)}_2^2.
\end{equation}
Depending on the type of problem and the availability of gradients, one may extend this by matching the gradients $\nabla L$ and $\nabla_\theta f_\phi$.
Modeling techniques for $f_\phi$ range from polynomials over radial basis functions and Kriging models to neural networks, and there is a distinction between global approximations of $L$ and ones that are valid only locally. Besides smaller models, the latter case may allow for error analysis through trust-region techniques \cite{Berkemeier2021}, at the cost of requiring additional intermittent evaluations of the original loss function $L$.

Not surprisingly, machine learning has found its entrance into this area of research as well, see \cite{Qu2021} for a recent overview. 
In the spirit of \eqref{eq:Surrogate}, deep surrogate models were suggested in \cite{Botache2024,Zhang2022}, and generative Kriging modeling was studied in \cite{Hussein2016}. Another generative modeling approach called GFlowNets was proposed in \cite{Jain2023}, and the usage of LLMs for solving MOPs was suggested in \cite{Liu2023}.
An alternative to surrogates for $L$ is to model the problem of hypervolume maximization by a deep neural network \cite{Zhang2023}.
Finally, besides supervised and generative modeling, there have been approaches using reinforcement learning \cite{Li2021,Zou2021} that suggest Pareto optimal points, and also to obtain the most efficient sample sites to build surrogates from when function evaluations are very expensive \cite{Chen2021}.

\section{Conclusion}\label{sec:Conclusion}
Multi-objective deep learning is constantly gaining attention, and we believe that the consideration of multiple conflicting criteria will become the new standard in the future, due to the ever-increasing complexity of modern-day tasks. And while researchers appear to have agreed on gradient-based approaches (MGDA) for multi-objective deep learning, there are many exciting questions for future research. 
\begin{itemize}
    \item interactive approaches where the training procedure interacts with a decision maker have not yet been studied,
    \item systematic usage for very high-dimensional problems with millions of parameters is still very scarce,
    \item challenging benchmark problems would help to foster research; in particular---to the best of our knowledge---there are no deep learning test problems where the Pareto front is non-convex,
    \item the massive trends of generative AI and large language models will likely play an important role as well---both in terms of solving multi-objective optimization problem and in using multi-objective optimization during their training.
\end{itemize}

\section*{Acknowledgments}
The authors acknowledge funding by the German Federal Ministry of Education and Research (BMBF) through the AI junior research group ``Multicriteria Machine Learning'' (Grant ID 01$|$S22064).

\bibliographystyle{IEEEtranS}
\bibliography{references}




\end{document}